\documentclass[runningheads]{llncs}
\usepackage[T1]{fontenc}
\usepackage{graphicx,verbatim}
\usepackage{multirow}
\usepackage{subcaption}
\usepackage{amsmath}
\usepackage{booktabs}
\usepackage{xcolor}

\begin{document}
%
% \title{Pose Estimation of Acetabular Hip Implants in Total Hip Arthroplasty Under Perspective Projection}
\title{2D/3D Registration of Acetabular Hip Implants Under Perspective Projection and Fully Differentiable Ellipse Fitting}
%
% \begin{comment}  %% Removed for anonymized MICCAI 2025 submission
\author{Yehyun Suh\inst{1,2,3} \and
J. Ryan Martin\inst{4} \and
Daniel Moyer\inst{1,2,3,*}}
\authorrunning{Y. Suh et al.}
\institute{Department of Computer Science, Vanderbilt University, Nashville TN 37235, USA \and
Vanderbilt Institute of Surgery and Engineering, Nashville TN 37235, USA \and
Vanderbilt Lab for Immersive AI Translation, Nashville TN 37235, USA \and
Department of Orthopaedic Surgery, Vanderbilt University Medical Center, Nashville TN 37232, USA\\
\email{\{yehyun.suh,daniel.moyer\}@vanderbilt.edu}}

% \end{comment}

% \author{Anonymized Authors}  %% Added for anonymized MICCAI 2025 submission
% \authorrunning{Anonymized Author et al.}
% \institute{Anonymized Affiliations \\
%     \email{email@anonymized.com}}

\maketitle              % typeset the header of the contribution
\begin{abstract}
This paper presents a novel method for estimating the orientation and the position of acetabular hip implants in total hip arthroplasty using full anterior-posterior hip fluoroscopy images. Our method accounts for distortions induced in the fluoroscope geometry, estimating acetabular component pose by creating a forward model of the perspective projection and implementing differentiable ellipse fitting for the similarity of our estimation from the ground truth.
This approach enables precise estimation of the implant’s rotation (anteversion, inclination) and the translation under the fluoroscope induced deformation. Experimental results from both numerically simulated and digitally reconstructed radiograph environments demonstrate high accuracy with minimal computational demands, offering enhanced precision and applicability in clinical and surgical settings.

\end{abstract}

\section{Introduction}
\label{sec:intro}
Total Hip Arthroplasty (THA) is a surgical procedure that replaces damaged hip joint cartilage and bone with artificial components. Complications and post-procedure negative outcomes have in part been ascribed to the orientation of the acetabular component (``hip joint cup'' or ``cup'') relative to the patient's natural position \cite{o2021impact}.

It is therefore important to provide intra-operative tracking and pose-estimation for the cup orientation. As the cups have a known and relatively simple geometry (effectively hemispheres of known radius), and have high radiodensity, fluoroscopy is the gold-standard option for localization and pose estimation. In the present work, we estimate the 3D pose of the implanted hemispheres given the observed ellipse in 2D fluoroscopy images.

This task would be analytically solvable using elementary mathematics for projections along standard Euclidean axes (orthographic projection). However, image intensifier fluoroscopes have a perspective projection geometry. Images are formed by projecting rays from the X-ray source (a point) to the collection panel (a plane), which results in deformations at coordinates away from the origin.
Current methods are largely focused on flat-plate or film collection \cite{Liaw_Hou_Yang_Wu_Fuh_2006b} cases, or stereotaxis (parallax) effects \cite{streck2023achieving}. The former have significant error when applied to hemispheres projected with this deformation. The latter requires multiple image collections from varying points of view, and error in C-arm (fluoroscope mount) motion or control induces further error.

2D/3D registration offers solution for estimating cup pose by generating a moving image from digitally reconstructed radiographs (DRR) \cite{gopalakrishnan2022fast,unberath2018deepdrr} and comparing it to a target image. While well-established, its practical application is limited by the computational inefficiency of repeatedly projecting the 3D implant model and the limited discriminative power of similarity calculations, which reduce registration accuracy. Additionally, requiring the exact 3D model—often unavailable in retrospective studies—further limits feasibility. However, this challenge of estimating a cup pose is a special case that can be addressed more efficiently. The method in this study overcomes these issues, enabling robust, real-world registration without exhaustive 3D modeling.

This paper proposes a method for estimating the acetabular hip implant pose in fluoroscopy images. Our contributions are:
\begin{itemize}
    \item pose estimation based on a perspective projection (cone-beam projection), avoiding the error in orthographic projection methods
    \item elimination of the need to project a full 3D model of the implant, significantly reducing computational costs
    \item error calculation using geometric descriptors of ellipses instead of landmark positions, solving a correspondence symmetry problem.
    \item a fully differentiable ellipse fitting process, enabling gradient-based optimization for precise pose estimation.
\end{itemize}

We demonstrate the effectiveness of our method by comparing it with orthographic projection for orientation estimation, and intensity-based and embedding-based 2D/3D registration methods for pose estimation. We also show the robustness in read-world data by implementing our method on cadaver data.

\section{Method}
Our objective is to recover the pose of the cup from intra-operative fluoroscopy. The pose is defined by five variables: anteversion/extra planar rotation $\theta$ (rotation into/out of the image plane), inclination/rotation within the image plane $\varphi$, in-plane translation $(k,l)$, and extra-planar translation $h$ of the cup. By convention these rotations are of the cup and not the imaging system/detector. The cups are assumed to be metal hemispheres with high radiodensity/image contrast, and annotated inner faces. Annotation of these structures in fluoroscopy can be performed automatically using e.g. a neural network \cite{ronneberger2015u,suh2023labelaugmentation}, which we demonstrate in several experiments but do not describe for brevity.

\begin{figure}[t]
    \centering
    \includegraphics[width=1\textwidth]{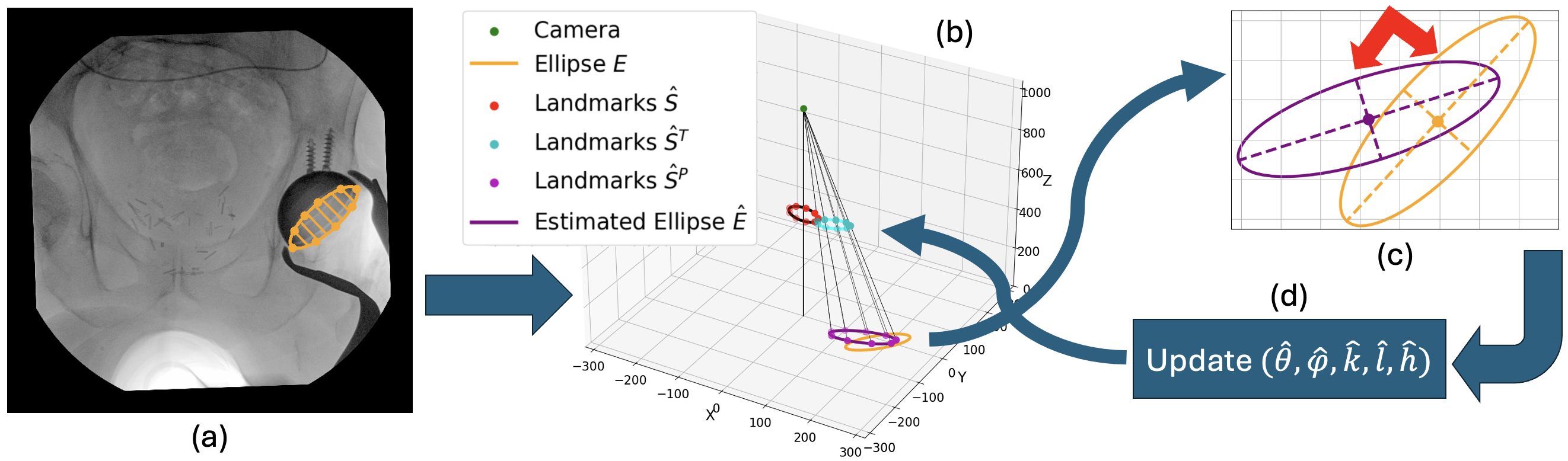}
    \caption{Registration pipeline. (a) \textbf{Segment} the ellipse (orange cross-line), \textbf{extract} landmarks $S^P$ (orange point) on the edge of the ellipse, and \textbf{calculate} ellipse $E$ (orange ellipsoidal-line) by fitting an ellipse on $S^P$. (b) Rotate and translate landmarks $\hat{S}$ (red) to $\hat{S}^T$ (light blue), \textbf{project} to obtain $\hat{S}^P$ (purple), and \textbf{calculate} ellipse $\hat{E}$ (purple line) by fitting an ellipse on $\hat{S}^P$. (c) Calculate the difference (red arrow) between $E$ and $\hat{E}$. (d) Update variable $(\hat\theta,\hat\varphi,\hat{k},\hat{l},\hat{h})$. Repeat process (b), (c), (d) until convergence.}
    \label{fig:method}
\end{figure}

Our method is composed of the following steps (shown graphically in Fig. \ref{fig:method}):
\begin{enumerate}
\item \label{est-step-1} Observe an ellipse $E$ using landmarks (denoted $S^P$) observed from the fluoroscopy by fitting an ellipse to it.
\item \label{est-step-2} Given a 3D pose of landmarks ($\hat{S}$), estimate the projected landmarks ($\hat{S}^P$) in the image plane and estimate the nominal projected ellipse ($\hat{E}$).
\item \label{est-step-3} Calculate error between the observed ellipse and the nominal projected ellipse parameters, then update the 3D pose via gradient descent (or another optimization method).
\item Repeat from Step \ref{est-step-2} until convergence.
\end{enumerate}
Notably, this method does not require correspondence between landmarks in $S^P$ and $\hat{S}^P$ in the image plane or in the estimated 3D pose landmarks $\hat{S}$.

\subsection{Ellipse Fitting from Landmarks}
Fitting the ellipse $E$ and $\hat{E}$ from landmarks $S^P$ and $\hat{S}^P$ (Fig \ref{fig:method}(a)) uses methods from Fitzgibbon et al. \cite{Fitzgibbon_Pilu_Fisher_1999} to form the least squares ellipse fit. This involves transforming the Euclidean coordinates of the landmarks into a six-dimensional space that parameterizes the implicit form of the ellipse equation:
\begin{align}
Ax^2+Bxy+Cy^2+Dx+Ey+F=0.
\end{align}
Each landmark coordinates $(x_i,y_i)$ in $S^P$ is transformed to a new vector:
\begin{align}
\left[ x_i, y_i \right] \mapsto \left[ x_i^2,x_iy_i,y_i^2,x_i,y_i,1 \right].
\end{align}
We concatenate these new vectors as rows in a data matrix $X_{ep}$. From this, the scatter matrix $M$ is calculated as:
\begin{align}
M = X_{\text{ep}}^T X_{\text{ep}}.
\end{align}
The generalized eigenvalue problem is solved using singular value decomposition (SVD) on the inverse of the scatter matrix, $M^{-1}$:
\begin{align}
U, S, V = \text{SVD}(M^{-1}).
\end{align}
The first column of U are the coefficients of the ellipse: $A,B,C,D,E,F=U_{:,1}$.
%\begin{align}
%    A,B,C,D,E,F=U_{:,1}.
%\end{align}
While the implicit parametrization uniquely describes each ellipse, error calculated in these parameters does not nicely correspond with intuitive notions of geometric error (e.g., Hausdorff differences between curves in the image plane). Thus, we convert coefficients to the standard parameterization. 
The center $(x,y)$ of the ellipse is:
\begin{align}
x = \frac{C \cdot D - \frac{B}{2} \cdot E}{2 \left( \left( \frac{B}{2} \right)^2 - A \cdot C \right)}, y = \frac{A \cdot E - \frac{B}{2} \cdot D}{2 \left( \left( \frac{B}{2} \right)^2 - A \cdot C \right)}.
\end{align}
%We define auxillary variables $m_{11}$, $m_{12}$, and $m_{22}$ as:
%\begin{align}
%m_{11} = \mu \cdot A, \quad m_{12} = \mu \cdot \frac{B}{2}, \quad m_{22} = \mu \cdot C, \nonumber\\
%\text{ where }\mu = \frac{1}{A \cdot x^2 + 2 \cdot \frac{B}{2} \cdot x \cdot y + C \cdot y^2 - F},
%\end{align}
%Defining $\mu$ as
For ease of notation, we can define a series of auxiliary variables:
\begin{align}
\mu = \frac{1}{A \cdot x^2 + 2 \cdot \frac{B}{2} \cdot x \cdot y + C \cdot y^2 - F},
\end{align}
and $m_{11} = \mu \cdot A$, $m_{12}= \mu \cdot \frac{B}{2}$, and $m_{22} = \mu \cdot C$. The semi-major and minor axes $a$ and $b$ are expressed in terms of these variables:
\begin{align}
a = \frac{1}{\frac{1}{2} \left( m_{11} + m_{22} + \sqrt{(m_{11} - m_{22})^2 + 4 \cdot m_{12}^2} \right)} \\
b = \frac{1}{\frac{1}{2} \left( m_{11} + m_{22} - \sqrt{(m_{11} - m_{22})^2 + 4 \cdot m_{12}^2} \right)}.
\end{align}
The orientation of the ellipse is calculated as:
\begin{align}
\alpha = \frac{1}{2} \cdot \text{atan2}\left( -2 \cdot \frac{B}{2}, C - A \right) \times \frac{180}{\pi},
\end{align}
%The output of applying fit ellipse calculation using the landmarks $S^P$ is the set of
and the parameters $\mathcal{E}=(x,y,a,b,\alpha)$ are the output of the process.

% Importantly, the entire ellipse fitting procedure is fully differentiable, as each step—from transforming landmark coordinates to computing the ellipse parameters—consists of differentiable operations. This property enables gradient-based optimization, allowing the estimated ellipse parameters to be refined iteratively within an optimization framework.
Importantly, the entire ellipse fitting is fully differentiable, as each step consists of differentiable operations. This enables gradient-based optimization, letting the parameters to be refined iteratively within an optimization framework.

\begin{figure}[t]
    \centering
    \begin{subfigure}{0.055\textwidth}
        \includegraphics[width=\textwidth]{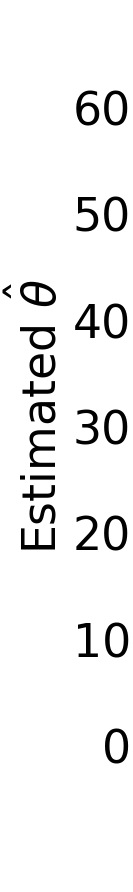}
    \end{subfigure}
    \begin{subfigure}{0.285\textwidth}
        \includegraphics[width=\textwidth]{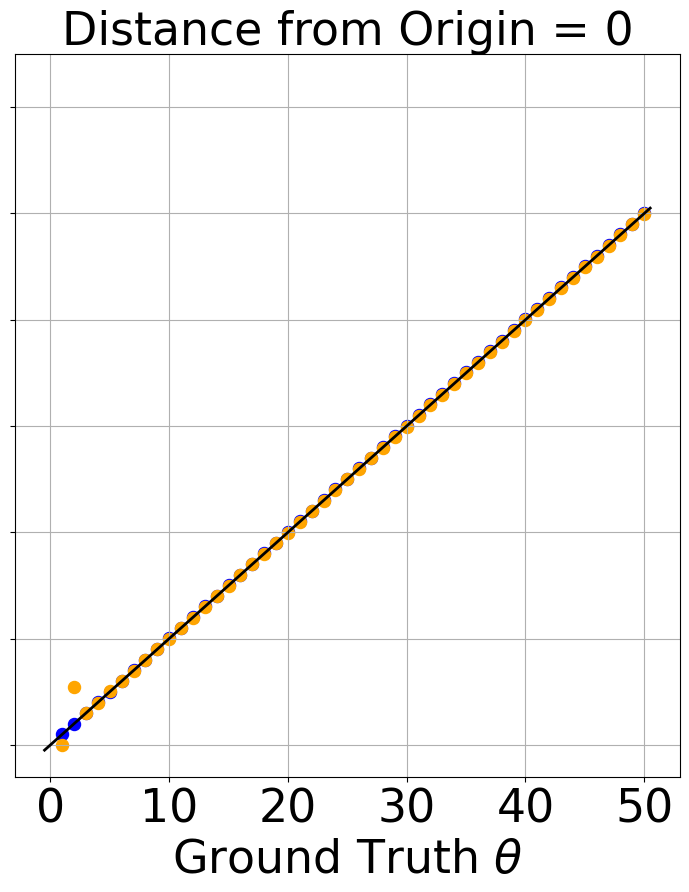}
    \end{subfigure}
    \begin{subfigure}{0.285\textwidth}
        \includegraphics[width=\textwidth]{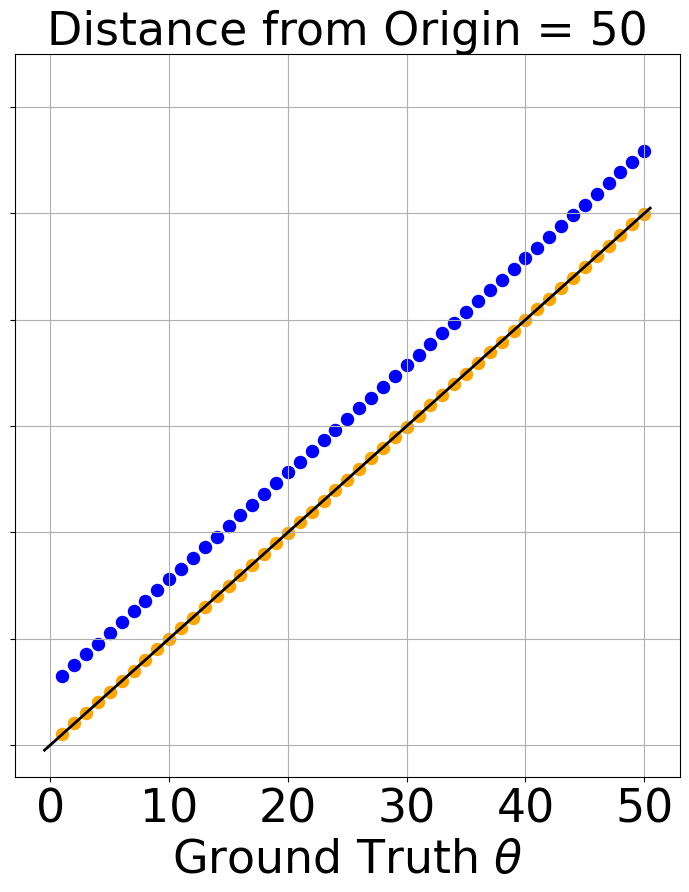}
    \end{subfigure}
    \begin{subfigure}{0.285\textwidth}
        \includegraphics[width=\textwidth]{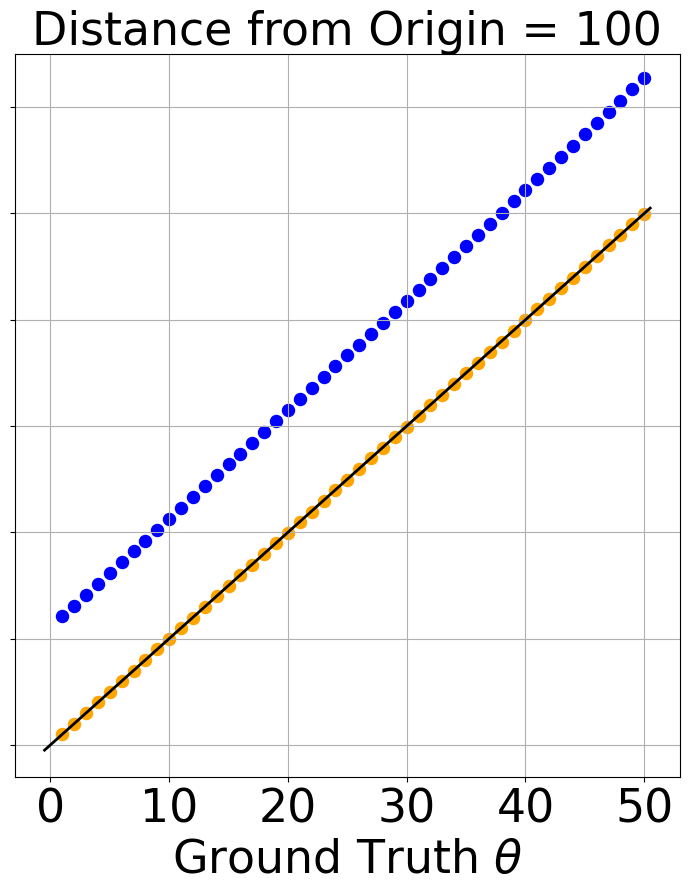}
    \end{subfigure}
    \begin{subfigure}{0.6\textwidth}
        \includegraphics[width=\textwidth]{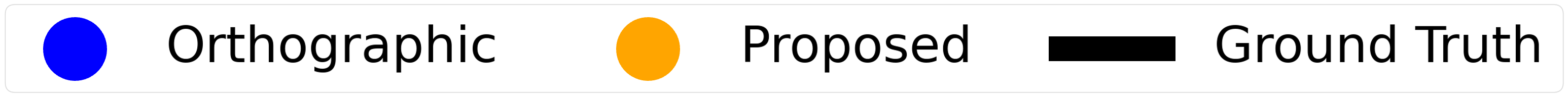}
    \end{subfigure}
    \caption{Comparison of $\hat{\theta}$ estimation on proposed and orthographic projection \cite{Liaw_Hou_Yang_Wu_Fuh_2006b} in the simulated environment. From left to right, experiments were conducted when distance from the origin to the object, $(k,l)$, increases from 0, 50, and to 100 mm, while other parameters were fixed. The peak at distance = 0 occurs as $\theta \rightarrow 0$, which causes numerical instability as the ellipse collapses to a line.}
    \label{fig:parallel vs proposed}
\end{figure}

\subsection{Forward Model: 3D-pose to the Image Plane}
% . To estimate the pose of the cup, standardly positioned landmarks $\hat{S}$ were used (Fig \ref{fig:method}(b)). These points are based on a circular coordinates with radius $r$ in a ``default'' position $(\theta,\varphi,k,l) = 0$

We start with landmark $\hat{S}$, which are points in a circle in a standard position, i.e., $(\theta,\varphi,k,l) = (0,0,0,0)$, at a mean distance $h$ along the extraplanar axis. The source is placed at height $H$ above the detection plane. Importantly, by convention, $\theta=0$ is orthogonal to the image plane (``zero ante/retroverson'' in THA), so $x_i = r \cos\left(\frac{2 \pi i}{n}\right), y_i = 0, z_i = h + r \sin\left(\frac{2 \pi i}{n}\right)$ for $n$ landmarks. The landmarks are rotated in 3D space by rotation matrix $R$:
\begin{align}
R(\theta,\varphi)=
\begin{pmatrix}
\cos\varphi & -\sin\varphi & 0 \\
\sin\varphi & \cos\varphi & 0 \\
0 & 0 & 1
\end{pmatrix}
\begin{pmatrix}
1 & 0 & 0 \\
0 & \cos\theta & -\sin\theta \\
0 & \sin\theta & \cos\theta
\end{pmatrix}.
\end{align}
By convention, we rotate by $\theta$ first, so as to avoid specifying the extra-planar rotation axis. The landmarks are then rotated and translated to $\hat{S}^T$, and projected to the 2D image plane:
\begin{align}
\hat{S}^T= R(\hat\theta,\hat\varphi)\hat{S} + \begin{pmatrix} \hat{k} \\ \hat{l} \\ 0 \end{pmatrix},
~~~~ ~~~~ 
\hat{S}^P=P(\hat{S}^T)=\frac{H}{H-\hat{S}^T_z}
\begin{pmatrix}
\hat{S}^T_x\\
\hat{S}^T_y\\
0
\end{pmatrix}
\end{align}
%$\hat{S}^T$ are projected on the 2D image plane by calculating the triangular ratio between the object and the image plane:
%\begin{align}
%\hat{S}^P=P(\hat{S}^T)=\frac{H}{H-\hat{S}^T_z}
%\begin{pmatrix}
%\hat{S}^T_x\\
%\hat{S}^T_y\\
%0
%\end{pmatrix}
%\end{align}
By using $\hat{S}^P$, we can fit another ellipse $\hat{E}$ with parameters $\hat{\mathcal{E}}=(\hat{x},\hat{y},\hat{a},\hat{b},\hat\alpha)$.

\subsection{Loss Calculation and Optimization}
We measure ellipse-to-ellipse distortion (Fig \ref{fig:method}(c)) using the mean squared error between parameter sets $\mathcal{E} = (x,y,a,b,\alpha)$ and the estimate $\hat{\mathcal{E}}$.
\begin{align}
\mathcal{L} = \frac{1}{N} \sum_{i=1}^N (\mathcal{E}_i - \hat{\mathcal{E}}_i)^2,
\end{align}
The error in the angular elements $\alpha$ are adjusted to account for angular periodicity, ensuring that discrepancies are correctly calculated by using $\min((\alpha - \hat{\alpha})^2, (180 - |\alpha - \hat{\alpha}|)^2)$. We then minimize the loss $\mathcal{L}$ by iteratively updating the parameters $(\hat\theta, \hat\varphi, \hat{k}, \hat{l}, \hat{h})$ using an analytical gradient computed through automatic differentiation (Fig \ref{fig:method}(d)), iterating until convergence.

\begin{table}[t]
\centering
% \caption{Quantitative results of pose estimation on simulated environment (top) and DRR of Implant (bottom). Evaluation are done under Hausdorff Distance (HD), total registration time, and errors. Mem. indicates the minimum GPU memories required during the registration process, when projected images size was $640\times 640$, where - represents cases that only used the CPU. Divergence (DIV) in the HD column indicates cases where the metric failed to provide a meaningful value due to extreme difference between the estimated and ground truth ellipses. This occurs because HD is highly sensitive to large spatial separations between two different ellipses, leading to divergence.}
\caption{Quantitative results of pose estimation on the simulated environment (top) and Implant CT (bottom) were evaluated using Hausdorff Distance (HD), registration time, and mean absolute rotation and translation errors. Mem. indicates the maximum GPU memory required for 2D/3D registration on $640\times 640$ images, with "-" for CPU-only cases. DIV marks where large in-plane spatial separations caused metric failure, as HD is highly sensitive to such differences.}
\label{tab:simulation_implant}
\begin{tabular}{@{}lccccccc@{}}
\toprule
Method & Mem. & HD & Time (s) & $\theta$ err ($^\circ$) & $\varphi$ err ($^\circ$) & $k, l$ err (mm) & $h$ err (mm) \\ \midrule
Orthographic \cite{Liaw_Hou_Yang_Wu_Fuh_2006b} & - & - & - & 5.46 & 2.18 & - & - \\
\textbf{Proposed} & - & 1.44 & 1.07 & 0.27 & 0.61 & 0.33 & 1.44 \\\bottomrule

Intensity \cite{gopalakrishnan2022fast} & 44 GB & 71.02 & 139.78 & 19.79 & 78.38 & 21.20 & 208.12 \\
Embedding \cite{gao2020ProST} & 42 GB & DIV & 286.25 & 18.21 & 38.20 & 71.02 & 207.78 \\
\textbf{Proposed} & - & 6.91 & 1.21 & 1.53 & 2.16 & 1.12 & 9.76 \\
Seg. + \textbf{Proposed} & 0.4 GB & 11.21 & 3.48 & 3.93 & 2.56 & 1.76 & 16.93 \\\bottomrule
\end{tabular}
\end{table}

\section{Experiments}
%\subsection{Environment for Extracting Landmark $S^P$}
Experiments were conducted in three distinct environments: Numerical Simulation, Implant CT, and Cadaver CT. For each of these environments, the pose parameters were uniformly sampled for each trial point: Anteversion ($\theta$) was sampled from the interval $(0,50)$ degrees (based on the ``safe zone'' in implant anteversion \cite{Abdel_Von_Roth_Jennings_Hanssen_Pagnano_2015,Lewinnek_Lewis_Tarr_Compere_Zimmerman_1978}), inclination ($\varphi$) from $(-90,90)$ degrees, in-plane translation ($k,l$) from $(-100,100)$ mm, and extra-planar translation ($h$) from $(100,520)$ mm. Projections onto the 2D plane were then simulated for the given pose parameters. All the experiments were conducted with source-to-detection distance $(H)$ set to $1040$ mm.

For optimization, post estimate variables were initialized to a ``first best guess'': $(\hat\theta_0,\hat\varphi_0,\hat{k}_0,\hat{l}_0,\hat{h}_0)=(25^\circ, 40^\circ,\beta E_x,\beta E_y, (1-\beta)H)$ where $E_x$ and $E_y$ is the center coordinates of the observed ellipse E, and $\beta$ is the ratio between the radius of the implant $r$ and major axis $E_a$ of the observed ellipse.

%The Hausdorff Distance (HD) between $E$ and $\hat{E}$ was calculated to assess fit quality for $\hat{E}$ \cite{rockafellar2009variational}.

For each of the environments, $S^{P}$ construction differs:
%\textbf{Numerical Simulation:}
In the \textbf{Numerical Simulation} environment, landmarks $S^P$ were directly projected to the image plane, i.e., no label or image noise was added for that experiment. %A ground truth landmark $S$ was generated and transformed by applying rotations and translations by $(\theta,\varphi,k,l,h)$, followed by projecting the landmarks into the 2D plane to obtain $S^P$.
For the \textbf{Implant CT} experiment environment, images were generated from CT of an implant outside of tissue using DRR methods. Two separate sub-experiments were conducted by extracting landmarks $S^P$ either by manually annotating landmarks in the CT and projecting them to the image plane, or, reflecting a more realistic scenario, training a U-Net \cite{ronneberger2015u} to segment the ellipsoidal area in the 2D image, the boundary of which are used as landmarks $S^P$.
%\textbf{Cadaver CT:}
In the \textbf{Cadaver CT} environment, 5 cadavers were given THA procedures and then imaged using CT. Simulated 2D imaging was then collected via DiffDRR, from which $S^P$ were annotated. Each cadaver has both Left and Right joint replacements.

We use the Numerical Simulation environment to validate the perspective projection over the orthographic projection methods \cite{Liaw_Hou_Yang_Wu_Fuh_2006b}. For the Implant CT environments we compare the performance of the proposed method to current 2D/3D image registration methods, either intensity-based \cite{gopalakrishnan2022fast} or embedding/feature based \cite{gao2020ProST}. We use the cadaver CT to demonstrate the viability of the proposed method in imaging conditions with real tissue.

%\textbf{Baselines:} For the simulated environment, we compare our method to orthographic projection \cite{Liaw_Hou_Yang_Wu_Fuh_2006b}. Also, for Implant CT, we compare 2 different 2D/3D registration methods where one compares the similarity of target and the moving image by the intensity of the images \cite{gopalakrishnan2022fast}, and the other one compares them by the embedding (feature) of the images \cite{gao2020ProST}. All the experiments were conducted with source-to-detection distance set to $1040$ mm.

\begin{figure}[t]
    \centering
    \begin{subfigure}{0.05\textwidth}
        \includegraphics[width=\textwidth]{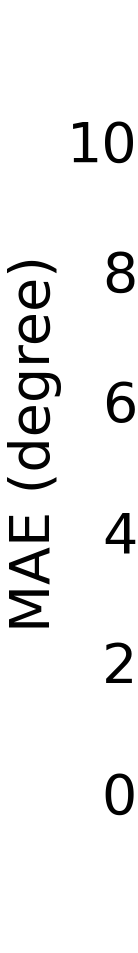}
    \end{subfigure}
    \begin{subfigure}{0.46\textwidth}
        \includegraphics[width=\textwidth]{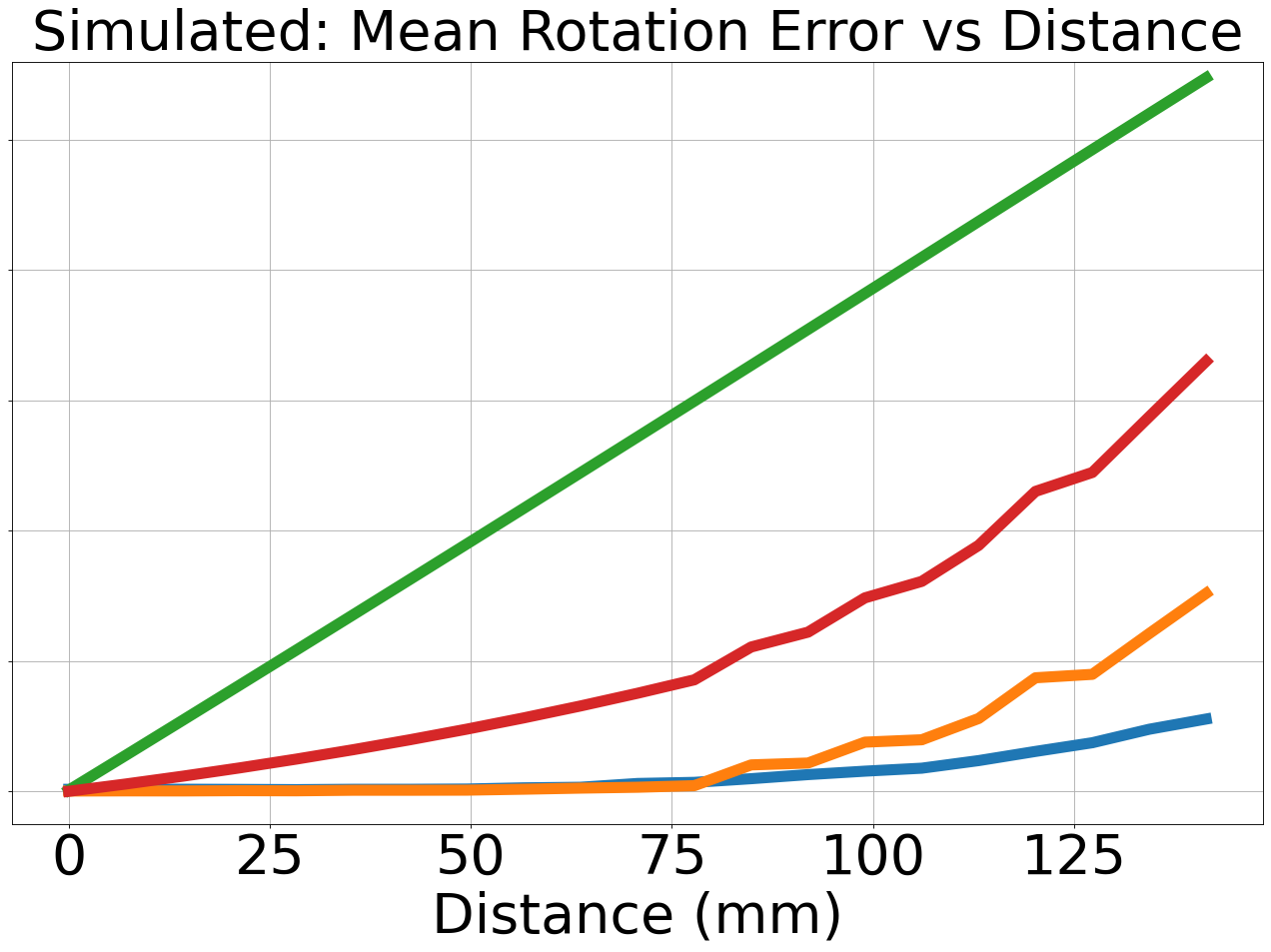}
    \end{subfigure}
    \begin{subfigure}{0.46\textwidth}
        \includegraphics[width=\textwidth]{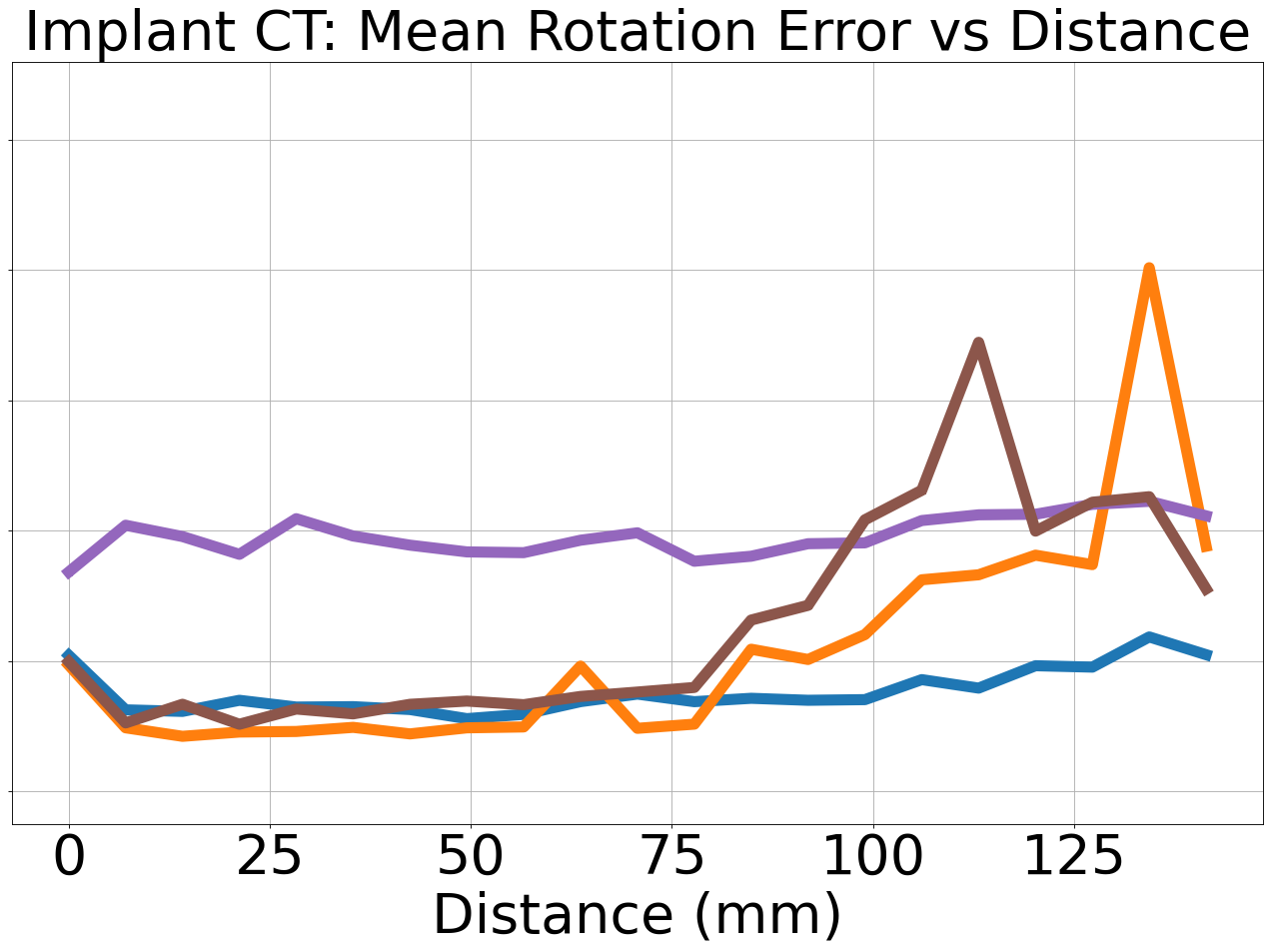}
    \end{subfigure} \\
    \begin{subfigure}{0.9\textwidth}
        \includegraphics[width=\textwidth]{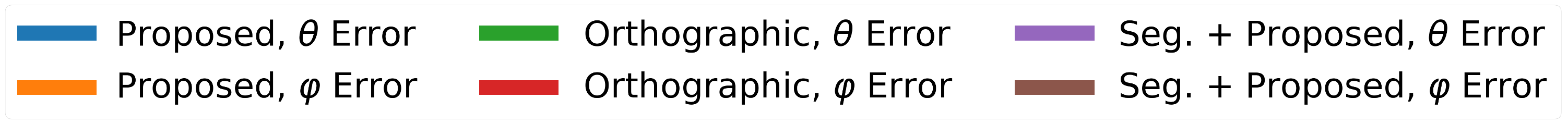}
    \end{subfigure}
    \caption{Mean absolute error (MAE) of $\theta$ and $\varphi$ as a function of the distance from the origin to the object, $(k,l)$. Experiments were conducted in the synthetic environment (left) and the DRR environment (right). $\theta$ and $\varphi$ indicate the experiments where each parameters were fixed. Results from intensity and embedding-based registration for implant CT was excluded due to high error.}
    \label{fig:average error}
\end{figure}

\section{Results}
% Synthetic Results
% \textbf{Simulation:} As stated in Table \ref{tab:simulation_implant}, our method applied to simulated scene showed average HD of 1.44 pixels where the process averaged 1.07 seconds for the whole registration. Our method outperformed the orthographic projection where the error for $\theta$ and $\varphi$ were $5.46^\circ$ and $2.18^\circ$. This is also shown in Fig \ref{fig:parallel vs proposed}, where the estimation's distortion from the ground truth linearly grew for orthographic projection. However, the proposed method was robust even though the implant got further away from the center.

\textbf{Numerical Simulation:} As shown in Table \ref{tab:simulation_implant}, our method achieved $\theta$ and $\varphi$ MAE of $0.27$ and $0.61$ degrees, respectively. In contrast, orthographic projection had $\theta$ and $\varphi$ errors of $5.46$ and $2.18$ degrees. Fig. \ref{fig:parallel vs proposed} further illustrates how orthographic projection caused increasing distortion, while the proposed method remained robust even as the implant moved away from the center.

% Implant CT Results
% \textbf{Implant CT:} As shown in Table \ref{tab:simulation_implant}, the proposed successfully registered the cup with an average HD of 6.91 pixels within 1.21 seconds. Other observed average errors were within $1.53^\circ$, $2.16^\circ$, $1.12$ mm, and $9.76$ mm for $(\hat\theta,\hat\varphi,(\hat{k},\hat{l}),\hat{h})$. When we obtained the ellipse from a U-Net model, overall HD and the estimation errors increased due to the minor mis-segmentation from the model. However, it was still faster and more accurate compared with other 2D/3D registration methods, which failed to register to the ground truth while consuming more up to 140 and 287 seconds for the whole registration process. The proposed method ran entirely on the CPU, requiring no GPU resources. Even with the segmentation model, it used only 0.4 GB of GPU memory, demonstrating its feasibility in the OR with minimal hardware requirements.

\begin{table}[t]
\centering
\caption{Quantative results on Cadaver CT using the proposed method showing the viability of the method in real surgical environments.}
\label{tab:cadaver}
\begin{tabular}{@{}lcccccccc@{}}
\toprule
Cadaver & Implant & HD & Time (s) & $\theta$ err ($^\circ$) & $\varphi$ err ($^\circ$) & $k, l$ err (mm) & $h$ err (mm) \\ \midrule

C 1 
& L & 2.18 & 1.08 & 0.92 & 0.22 & 0.48 & 3.33 \\
& R & 6.26 & 1.09 & 1.23 & 0.31 & 1.62 & 8.77 \\ \midrule

C 2 
& L & 3.59 & 1.09 & 0.38 & 0.64 & 0.73 & 6.07 \\
& R & 0.24 & 1.12 & 0.52 & 1.37 & 0.65 & 1.42 \\ \midrule

C 3 
& L & 2.58 & 1.08 & 0.25 & 0.97 & 0.58 & 4.63 \\
& R & 2.72 & 1.08 & 0.39 & 0.43 & 1.55 & 5.61 \\ \midrule

C 4 
& L & 3.67 & 2.86 & 0.05 & 1.51 & 1.26 & 8.48 \\
& R & 2.22 & 1.08 & 0.28 & 0.84 & 1.56 & 5.86 \\ \midrule

C 5 
& L & 6.94 & 1.07 & 0.46 & 0.31 & 0.56 & 6.28 \\
& R & 0.70 & 2.18 & 0.22 & 0.87 & 0.84 & 0.15 \\ \midrule

Mean & & 3.11 & 1.38 & 0.47 & 0.75 & 0.99 & 5.07 \\

\bottomrule
\end{tabular}
\end{table}

% \begin{table}[t]
% \centering
% \caption{Quantative results on Cadaver CT environment using the proposed method showing the viability of the method in real surgical environments.}
% \label{tab:cadaver}
% \begin{tabular}{@{}lcccccccc@{}}
% \toprule
% Cadaver & Implant & HD & Time (s) & $\theta$ err ($^\circ$) & $\varphi$ err ($^\circ$) & $k, l$ err (mm) & $h$ err (mm) \\ \midrule

% C 1 & L & 4.49 & 1.07 & 0.98 & 0.19 & 0.76 & 5.81 \\
% & R & 2.11 & 1.06 & 0.30 & 0.11 & 0.27 & 3.77 \\ \midrule

% C 2 & L & 4.66 & 1.08 & 1.04 & 0.21 & 1.11 & 7.40 \\
% & R & 0.87 & 1.09 & 0.01 & 0.01 & 0.27 & 2.23 \\ \midrule

% C 3 & L & 0.61 & 1.08 & 0.02 & 0.01 & 0.22 & 2.27 \\
% & R & 3.13 & 1.07 & 0.99 & 0.11 & 1.22 & 5.97 \\ \midrule

% C 4 & L & 1.17 & 1.03 & 0.00 & 0.02 & 0.50 & 3.51 \\
% & R & 0.96 & 1.08 & 0.02 & 0.02 & 0.33 & 2.96 \\ \midrule

% C 5 & L & 2.08 & 1.06 & 0.15 & 0.00 & 0.13 & 2.42 \\
% & R & 0.83 & 1.03 & 0.00 & 0.05 & 0.36 & 2.57 \\ \midrule

% Mean & & 2.09 & 1.07 & 0.35 & 0.07 & 0.52 & 3.89 \\

% \bottomrule
% \end{tabular}
% \end{table}

\textbf{Implant CT:} As shown in Table \ref{tab:simulation_implant}, the proposed method successfully registered the cup with an average HD of 6.91 pixels in just 1.21 seconds. Other errors remained within $1.53$ and $2.16$ degrees, $1.12$ and $9.76$ mm for $(\hat\theta,\hat\varphi,(\hat{k},\hat{l}),\hat{h})$. Incorporating a segmentation model led to slightly higher errors due to minor mis-segmentations, but still outperformed other 2D/3D registration methods which failed to register correctly while consuming up to 287 seconds. The robustness of our method is also shown in Fig. \ref{fig:average error}, where estimation errors remain consistently low across varying distances. Moreover, this method operates entirely on the CPU, and even with segmentation, it only requires 0.4 GB of GPU memory, making it practical in the operating room with minimal hardware requirements.

\textbf{Cadaver CT:} As shown in Table \ref{tab:cadaver}, our method successfully registered to implants in cadaver data, achieving a mean HD of 3.17 pixels with an average registration time of 1.38 seconds. The improved performance on cadaver data compared to implant images is due to the clear visibility of ellipses in the DRRs of implants in the cadaver. In contrast, the implant CT environment include cases where ellipses are barely visible, leading to higher HD values and errors.

% \section{Discussion}
% This work can be thought of as an advantageous and important special case of the general inverse problem in projective geometry. Solutions to the general case such as Uneri et al. \cite{Uneri_2015_KC_3D_2D_Reg} require iterative forward estimation of the image; for this particular special case we can solve this forward operation analytically, producing a much more computationally efficient solution. Landmark registration methods are also used in place of full image projection \cite{Grupp_Unberath_Gao_Hegeman_Murphy_Alexander_Otake_McArthur_Armand_Taylor_2020}. These require correspondence between landmarks; while artificial correspondences could be imposed as in iterative closest point (ICP) methods \cite{besl1992method}, we avoid these complications by directly using the ellipse geometry.

% \section{Conclusion}
% The proposed method for hip implant orientation estimation in THA demonstrates high precision and computational efficiency by utilizing perspective projection and fit ellipse calculations from synthetic 3D landmarks. This approach reduces computational load by avoiding the projection of a full 3D image, achieving processing capabilities within seconds while addressing projection distortions, especially as the implant moves away from the image center. The method's high accuracy in the test cases proves the robustness of the method in diverse environments.

\section{Discussion and Conclusion}  
This work presents an efficient and precise method for hip implant orientation estimation in THA, proving it as an advantageous and important special case of the broader inverse problem in projective geometry. While general solutions, such as those by Uneri et al. \cite{Uneri_2015_KC_3D_2D_Reg}, require iterative forward estimations, our approach analytically solves the forward operation, significantly improving computational efficiency. Similarly, landmark registration methods \cite{Grupp_Unberath_Gao_Hegeman_Murphy_Alexander_Otake_McArthur_Armand_Taylor_2020} often rely on explicit correspondences between landmarks, which we solve by applying ellipse geometry instead of imposing artificial correspondences as in ICP methods \cite{besl1992method}. 

By integrating perspective projection and ellipse fitting from landmarks, our method reduces computational inefficiency, achieving near real-time processing while mitigating projection distortions, particularly as the implant gets further from the image center. The high accuracy demonstrated in test cases on cadavers confirm the robustness of this approach in real surgical conditions. Furthermore, the differentiability of the ellipse fitting process allows for its integration into learning-based frameworks, such as deep neural networks. This facilitates the incorporation of data-driven refinement strategies, potentially enhancing pose estimation accuracy through end-to-end optimization.

% \begin{comment}  %% removed for anonymized MICCAI 2025 submission.

\begin{credits}
\subsubsection{\ackname} This work was supported in part by NSF 2321684 and a VISE Seed Grant.

\subsubsection{\discintname} The authors have no competing interests to declare that are relevant to the content of this article.
\end{credits}

% \end{comment}
\bibliographystyle{splncs04}
\bibliography{ref}

\begin{thebibliography}{10}
\providecommand{\url}[1]{\texttt{#1}}
\providecommand{\urlprefix}{URL }
\providecommand{\doi}[1]{https://doi.org/#1}

\bibitem{Abdel_Von_Roth_Jennings_Hanssen_Pagnano_2015}
Abdel, M.P., Von~Roth, P., Jennings, M.T., Hanssen, A.D., Pagnano, M.W.: What safe zone? the vast majority of dislocated thas are within the lewinnek safe zone for acetabular component position. Clinical Orthopaedics and Related Research  \textbf{474}(2),  386–391 (Jul 2015). \doi{10.1007/s11999-015-4432-5}, \url{https://pmc.ncbi.nlm.nih.gov/articles/PMC4709312}

\bibitem{besl1992method}
Besl, P.J., McKay, N.D.: Method for registration of 3-d shapes. In: Sensor fusion IV: control paradigms and data structures. vol.~1611, pp. 586--606. Spie (1992)

\bibitem{Fitzgibbon_Pilu_Fisher_1999}
Fitzgibbon, A., Pilu, M., Fisher, R.: Direct least square fitting of ellipses. IEEE Transactions on Pattern Analysis and Machine Intelligence  \textbf{21}(5),  476–480 (May 1999). \doi{10.1109/34.765658}, \url{https://doi.org/10.1109/34.765658}

\bibitem{gao2020ProST}
Gao, C., Liu, X., Gu, W., Killeen, B., Armand, M., Taylor, R., Unberath, M.: Generalizing spatial transformers to projective geometry with applications to 2d/3d registration. In: Medical Image Computing and Computer Assisted Intervention--MICCAI 2020: 23rd International Conference, Lima, Peru, October 4--8, 2020, Proceedings, Part III 23. pp. 329--339. Springer (2020)

\bibitem{gopalakrishnan2022fast}
Gopalakrishnan, V., Golland, P.: Fast auto-differentiable digitally reconstructed radiographs for solving inverse problems in intraoperative imaging. In: Workshop on Clinical Image-Based Procedures. pp. 1--11. Springer (2022)

\bibitem{Grupp_Unberath_Gao_Hegeman_Murphy_Alexander_Otake_McArthur_Armand_Taylor_2020}
Grupp, R.B., Unberath, M., Gao, C., Hegeman, R.A., Murphy, R.J., Alexander, C.P., Otake, Y., McArthur, B.A., Armand, M., Taylor, R.H.: Automatic annotation of hip anatomy in fluoroscopy for robust and efficient 2d/3d registration. International Journal of Computer Assisted Radiology and Surgery  \textbf{15}(5),  759–769 (Apr 2020). \doi{10.1007/s11548-020-02162-7}, \url{https://doi.org/10.1007/s11548-020-02162-7}

\bibitem{Lewinnek_Lewis_Tarr_Compere_Zimmerman_1978}
Lewinnek, G.E., Lewis, J.L., Tarr, R., Compere, C.L., Zimmerman, J.R.: Dislocations after total hip-replacement arthroplasties. Journal of Bone and Joint Surgery  \textbf{60}(2),  217–220 (Mar 1978). \doi{10.2106/00004623-197860020-00014}, \url{https://doi.org/10.2106/00004623-197860020-00014}

\bibitem{Liaw_Hou_Yang_Wu_Fuh_2006b}
Liaw, C.K., Hou, S.M., Yang, R.S., Wu, T.Y., Fuh, C.S.: A new tool for measuring cup orientation in total hip arthroplasties from plain radiographs. Clinical Orthopaedics and Related Research  \textbf{451},  134–139 (Jun 2006). \doi{10.1097/01.blo.0000223988.41776.fa}, \url{https://doi.org/10.1097/01.blo.0000223988.41776.fa}

\bibitem{o2021impact}
O'Connor, P.B., Thompson, M.T., Esposito, C.I., Poli, N., McGree, J., Donnelly, T., Donnelly, W.: The impact of functional combined anteversion on hip range of motion: a new optimal zone to reduce risk of impingement in total hip arthroplasty. Bone \& Joint Open  \textbf{2}(10),  834--841 (2021)

\bibitem{ronneberger2015u}
Ronneberger, O., Fischer, P., Brox, T.: U-net: Convolutional networks for biomedical image segmentation. In: Medical image computing and computer-assisted intervention--MICCAI 2015: 18th international conference, Munich, Germany, October 5-9, 2015, proceedings, part III 18. pp. 234--241. Springer (2015)

\bibitem{streck2023achieving}
Streck, L.E., Boettner, F.: Achieving precise cup positioning in direct anterior total hip arthroplasty: a narrative review. Medicina  \textbf{59}(2), ~271 (2023)

\bibitem{suh2023labelaugmentation}
Suh, Y., Chan, P., Martin, J.R., Moyer, D.: Label augmentation method for medical landmark detection in hip radiograph images (2023), \url{https://arxiv.org/abs/2309.16066}

\bibitem{unberath2018deepdrr}
Unberath, M., Zaech, J.N., Lee, S.C., Bier, B., Fotouhi, J., Armand, M., Navab, N.: Deepdrr--a catalyst for machine learning in fluoroscopy-guided procedures. In: Medical Image Computing and Computer Assisted Intervention--MICCAI 2018: 21st International Conference, Granada, Spain, September 16-20, 2018, Proceedings, Part IV 11. pp. 98--106. Springer (2018)

\bibitem{Uneri_2015_KC_3D_2D_Reg}
Uneri, A., Stayman, J.W., Silva, T.D., Wang, A.S., Kleinszig, G., Vogt, S., M.D., A.J.K., M.D., J.P.W., M.D., Z.L.G., Siewerdsen, J.H.: {Known-component 3D-2D registration for image guidance and quality assurance in spine surgery pedicle screw placement}. In: III, R.J.W., Yaniv, Z.R. (eds.) Medical Imaging 2015: Image-Guided Procedures, Robotic Interventions, and Modeling. vol.~9415, p. 94151F. International Society for Optics and Photonics, SPIE (2015). \doi{10.1117/12.2082210}, \url{https://doi.org/10.1117/12.2082210}

\end{thebibliography}
\end{document}